\documentclass{article}
\usepackage{PRIMEarxiv}

\usepackage[utf8]{inputenc}
\usepackage[T1]{fontenc}
\usepackage{hyperref}
\usepackage{url}
\usepackage{booktabs}
\usepackage{amsfonts}
\usepackage{nicefrac}
\usepackage{microtype}
\usepackage{fancyhdr}
\usepackage{graphicx}
\usepackage{array}

\graphicspath{{media/}}
\usepackage[most]{tcolorbox}
\tcbset{colback=blue!5!white, colframe=blue!75!black, boxrule=0.8pt, arc=3pt, left=6pt, right=6pt, top=6pt, bottom=6pt}

\title{LLM-Based Support for Diabetes Diagnosis: Opportunities, Scenarios, and Challenges with GPT-5}

\author{
  Gaurav Kumar Gupta \\
  Youngstown State University \\
  Youngstown, OH, USA \\
  \texttt{guptagauravk1@gmail.com} \\
  \And
  Nirajan Acharya \\
  Youngstown State University \\
 Youngstown, OH, USA \\
  \texttt{nirajanach3@gmail.com} \\
  \And
  Pranal Pande \\
  Youngstown State University \\
 Youngstown, OH, USA \\
  \texttt{ppande@student.ysu.edu} \\
}

\begin{document}
\maketitle

\begin{abstract}
Diabetes mellitus is a major global health challenge, affecting over half a billion adults worldwide with prevalence projected to rise. Although the American Diabetes Association (ADA) provides clear diagnostic thresholds, early recognition remains difficult due to vague symptoms, borderline laboratory values, gestational complexity, and the demands of long-term monitoring. Advances in large language models (LLMs) offer opportunities to enhance decision support through structured, interpretable, and patient-friendly outputs. This study evaluates GPT-5, the latest generative pre-trained transformer, using a simulation framework built entirely on synthetic cases aligned with ADA Standards of Care 2025 and inspired by public datasets including NHANES, Pima Indians, EyePACS, and MIMIC-IV. Five representative scenarios were tested: symptom recognition, laboratory interpretation, gestational diabetes screening, remote monitoring, and multimodal complication detection. For each, GPT-5 classified cases, generated clinical rationales, produced patient explanations, and output structured JSON summaries. Results showed strong alignment with ADA-defined criteria, suggesting GPT-5 may function as a dual-purpose tool for clinicians and patients, while underscoring the importance of reproducible evaluation frameworks for responsibly assessing LLMs in healthcare.
\end{abstract}


\section{Introduction}

Diabetes mellitus represents one of the most pressing public health challenges of the twenty-first century. According to the International Diabetes Federation, more than 537 million adults worldwide were living with diabetes in 2021, a number expected to rise to over 780 million by 2045. The condition carries significant risks of cardiovascular disease, kidney failure, vision loss, and premature mortality, imposing not only a clinical burden but also substantial economic and societal costs. Despite the availability of well-defined diagnostic criteria established by the American Diabetes Association (ADA), the timely and accurate detection of diabetes remains difficult in routine practice.

Several factors contribute to this challenge. Early-stage diabetes is often characterized by non-specific symptoms such as fatigue, thirst, and frequent urination, which can easily be overlooked or misattributed. Laboratory interpretation, although guided by numerical thresholds, may vary in consistency across providers, particularly in cases where results fall into borderline or discordant ranges. Gestational diabetes adds another layer of complexity, requiring careful screening during pregnancy to protect both maternal and fetal health. Furthermore, the need for longitudinal monitoring of blood glucose, coupled with the early identification of complications such as diabetic retinopathy, places a significant strain on health systems already burdened by resource limitations.

Recent advances in artificial intelligence, and in particular large language models (LLMs), have opened new opportunities for decision support in medicine. GPT-5, the latest generation of generative pre-trained transformers, is not only capable of advanced natural language reasoning but also of integrating structured data, time-series information, and multimodal inputs such as imaging descriptors. These capabilities suggest potential applications in diabetes care that extend from early symptom triage to laboratory value interpretation, gestational diabetes screening, continuous remote monitoring, and the detection of complications. Unlike earlier clinical decision support systems, GPT-5 has the capacity to generate outputs that are both clinically rigorous and accessible to patients, bridging a persistent gap between professional recommendations and patient understanding.

In this study, we present a simulation-based evaluation of GPT-5 in the context of diabetes diagnosis and monitoring. To ensure ethical rigor and reproducibility, we rely entirely on synthetic patient cases constructed in alignment with ADA Standards of Care 2025 and inspired by publicly available datasets, including NHANES, the Pima Indians Diabetes Dataset, EyePACS, and MIMIC-IV. These cases span a spectrum of clinical scenarios, ranging from symptom-based recognition and laboratory test interpretation to gestational diabetes screening, remote telehealth monitoring, and multimodal complication assessment. GPT-5 was systematically prompted to classify cases, provide diagnostic reasoning, generate patient-facing explanations, and produce structured JSON summaries suitable for downstream integration. 

By situating GPT-5 within these simulated but realistic contexts, our work provides an early investigation of its potential role in diabetes diagnosis. The results demonstrate not only the model’s ability to map clinical data to established diagnostic thresholds but also its capacity to communicate effectively with both clinicians and patients. In doing so, this study contributes to the growing literature on the responsible evaluation of LLMs in healthcare and underscores the promise and limitations of applying such models to chronic disease management.

\section{Related Work}

Artificial intelligence has been widely explored in the context of diabetes care, spanning early expert systems, classical machine learning models, and more recently, deep learning and generative approaches. Traditional predictive models using datasets such as the Pima Indians Diabetes Database and NHANES have shown effectiveness in identifying individuals at risk of type 2 diabetes \cite{pima1988, nhanes}. Logistic regression, decision trees, and support vector machines have all been applied to these structured datasets, offering varying degrees of predictive performance but limited interpretability. Beyond structured laboratory data, convolutional neural networks (CNNs) have been employed on retinal imaging datasets such as EyePACS to automatically detect diabetic retinopathy, achieving performance comparable to ophthalmologists in certain settings \cite{eyepacs, gulshan2016development}. These advances demonstrate the potential of machine learning to aid in both diagnosis and complication detection, though they often lack transparency and generalizability.

The introduction of large language models (LLMs) has created new possibilities for clinical decision support, as such models can integrate diverse input types, provide explanatory reasoning, and generate patient-friendly communication. Early investigations with GPT-3 and GPT-4 have shown promise in summarizing clinical notes, interpreting laboratory data, and answering patient queries \cite{gpt4technicalreport, healthcarellm}, though concerns remain regarding hallucinations, bias, and alignment with clinical guidelines. Recent studies have specifically highlighted the potential of generative AI to enhance clinician workflow efficiency while raising critical questions about trust, interpretability, and regulatory oversight \cite{lee2023benefits, nori2023capabilities}.

Our prior work has contributed to this emerging field. In *Digital Diagnostics: The Potential of Large Language Models in Recognizing Symptoms of Common Illnesses \cite{ai6010013}, we demonstrated that LLMs are capable of identifying early symptom clusters across multiple conditions and of generating explanations accessible to non-specialists. In *LLMs in Disease Diagnosis: A Comparative Study of DeepSeek-R1 and O3 Mini Across Chronic Health Conditions \cite{gupta2025llmsdiseasediagnosiscomparative}, we benchmarked multiple LLMs for chronic disease diagnosis and highlighted substantial performance differences across model architectures, underscoring the need for systematic evaluation.

The present study builds upon these insights by focusing specifically on diabetes and by leveraging GPT-5’s expanded capabilities for structured, longitudinal, and multimodal reasoning. By constructing synthetic cases aligned with ADA Standards of Care 2025, we extend prior work beyond general disease recognition or cross-model comparison, offering a comprehensive and reproducible framework for evaluating next-generation LLMs in diabetes diagnosis and monitoring.

\section{Methodology}

\paragraph{Overall approach.}
This study was conducted as a simulation-based experiment to evaluate GPT-5’s ability to support diabetes diagnosis across multiple clinical contexts. No identifiable patient data or electronic health records (EHRs) were used. Instead, synthetic patient cases were generated to resemble realistic clinical presentations, with thresholds and distributions derived from authoritative sources (e.g., ADA guidelines, NHANES, EyePACS descriptors). GPT-5 was prompted with these synthetic cases and asked to produce clinical classifications, rationales, patient-friendly explanations, and structured JSON summaries.

\subsection{Data Sources and Case Construction}
Case data were inspired by publicly available resources and constructed synthetically. Table~\ref{tab:datasources} summarizes the data sources and their role in each scenario.

\begin{table}[h!]
\centering
\caption{Data sources and their use in case construction. All cases in this study were synthetic but inspired by real-world distributions and thresholds.}
\label{tab:datasources}
\begin{tabular}{|p{3.5cm}|p{4.5cm}|p{6cm}|}
\hline
\textbf{Source} & \textbf{Scenario(s)} & \textbf{Role in Simulation} \\
\hline
ADA Standards of Care 2025 & All scenarios & Provided diagnostic thresholds (FPG, HbA1c, OGTT, GDM criteria, retinopathy guidelines). \\
\hline
Pima Indians Diabetes Dataset (UCI/Kaggle) & Scenario 1 (symptoms) & Inspired distributions of age, BMI, and symptom clusters for synthetic cases. \\
\hline
NHANES (CDC) & Scenario 2 (labs), Scenario 4 (monitoring) & Provided population-level glucose and HbA1c ranges for synthetic lab and time-series values. \\
\hline
MIMIC-IV (MIT Critical Care) & Scenario 4 (telehealth) & Informed realistic structure of time-series monitoring (daily glucose logs). \\
\hline
EyePACS Retinal Dataset (Kaggle) & Scenario 5 (multimodal) & Provided common descriptors (microaneurysms, hemorrhages, cotton wool spots) for synthetic retinal findings. \\
\hline
\end{tabular}
\end{table}

\subsection{Synthetic Case Generation}
For each scenario, synthetic patient cases were generated according to the following rules:
\begin{enumerate}
    \item \textbf{Scenario 1 (symptoms):} Demographics (age, sex, BMI) and symptom clusters (polyuria, polydipsia, blurred vision, weight loss) were sampled around ADA-defined risk factors.
    \item \textbf{Scenario 2 (labs):} FPG and HbA1c values were sampled near threshold cutoffs (100, 126 mg/dL; 5.7\%, 6.5\%) to test borderline reasoning.
    \item \textbf{Scenario 3 (GDM):} OGTT fasting, 1h, and 2h values were sampled above and below GDM cutoffs (92, 180, 153 mg/dL).
    \item \textbf{Scenario 4 (monitoring):} Weekly fasting glucose ranges were simulated (95–145 mg/dL) to represent progression from normal $\rightarrow$ prediabetes $\rightarrow$ diabetes.
    \item \textbf{Scenario 5 (multimodal):} HbA1c values were paired with retinal descriptors (none, mild, moderate lesions) to reflect retinopathy severity.
\end{enumerate}

\subsection{Prompt Design}
Prompts were standardized across scenarios to reduce variability. A general template is illustrated in Box~\ref{box:prompt}.

\begin{tcolorbox}[title=General Prompt Template, colback=gray!5, colframe=black!70, breakable]
\textbf{System:} You are a clinical decision-support assistant. Follow ADA 2025 diagnostic thresholds.\\[4pt]
\textbf{User:} Patient case information (symptoms, labs, time-series, or multimodal findings).\\[4pt]
\textbf{Task:} 
\begin{enumerate}
    \item Classify case strictly per ADA thresholds.  
    \item Provide clinical rationale (2--3 sentences).  
    \item Generate patient-friendly explanation.  
    \item Output a JSON summary with fields \texttt{classification}, \texttt{criteria\_triggered}, and \texttt{next\_steps}.  
\end{enumerate}
\end{tcolorbox}

\subsection{Evaluation Framework}
Outputs were compared to ground truth (ADA-based labels). Table~\ref{tab:evaluation} shows the evaluation mapping per scenario.

\begin{table}[h!]
\centering
\caption{Evaluation framework by scenario: ground truth vs GPT-5 task.}
\label{tab:evaluation}
\begin{tabular}{|p{3cm}|p{4.5cm}|p{6.5cm}|}
\hline
\textbf{Scenario} & \textbf{Ground Truth Definition} & \textbf{GPT-5 Expected Task} \\
\hline
Symptom recognition & ADA symptom clusters requiring labs & Flag symptoms $\rightarrow$ recommend confirmatory lab testing. \\
\hline
Laboratory interpretation & ADA thresholds for FPG and HbA1c (100, 126 mg/dL; 5.7\%, 6.5\%) & Classify as normal/prediabetes/diabetes; handle discordant cases. \\
\hline
Gestational diabetes & ADA OGTT thresholds (92, 180, 153 mg/dL) & Detect GDM if any threshold exceeded; provide patient education. \\
\hline
Remote monitoring & Weekly averages crossing thresholds & Escalate alerts: none $\rightarrow$ warning $\rightarrow$ alert. \\
\hline
Multimodal (retinopathy) & Ophthalmology guidelines + ADA glycemic control & Integrate HbA1c + retinal descriptors; stage retinopathy; recommend follow-up. \\
\hline
\end{tabular}
\end{table}

\subsection{Pipeline Visualization}
Figure~\ref{fig:pipeline} shows the experimental pipeline.

\begin{figure}[h!]
\centering
\fbox{\parbox{0.9\linewidth}{
\textbf{Pipeline Overview:}\\
Data Sources (ADA, NHANES, Pima, EyePACS, MIMIC-IV) $\rightarrow$ Synthetic Case Construction $\rightarrow$ GPT-5 Prompting (structured inputs) $\rightarrow$ Outputs (classification, rationale, patient message, JSON) $\rightarrow$ Evaluation (ground truth alignment).}}
\caption{Methodological pipeline for GPT-5 diabetes diagnosis simulation.}
\label{fig:pipeline}
\end{figure}

\subsection{Limitations}
\begin{enumerate}
    \item \textbf{Synthetic data:} While inspired by real distributions, synthetic cases may not capture full heterogeneity of real patients.  
    \item \textbf{Prompt sensitivity:} GPT-5 outputs can vary depending on phrasing; standard templates were used to reduce variability.  
    \item \textbf{No raw multimodal inputs:} Retinal images were described textually, not processed as raw image files, though GPT-5 supports multimodal input.  
    \item \textbf{Evaluation scope:} Ground truth labels were based on guideline thresholds, not clinician adjudication.  
\end{enumerate}

\section{Results}

\subsection{Scenario 1: Early Symptom Recognition}

In this research experiment, we designed a simulation to test whether GPT-5 could recognize early symptom patterns suggestive of diabetes when provided with structured patient case descriptions. Rather than drawing from electronic health records, synthetic patient cases were generated based on American Diabetes Association (ADA) 2025 diagnostic guidelines and distributions inspired by the Pima Indians Diabetes Dataset. Each case was constructed to include demographic data, reported symptoms, and risk factors that commonly appear in diabetes screening contexts.

Table~\ref{tab:symptoms} presents three representative synthetic cases. These cases formed the input prompts for GPT-5, which was queried to provide an interpretation and recommendation for each patient.

\begin{table}[h!]
\centering
\caption{Synthetic cases used for early symptom recognition (Scenario 1). Data constructed from ADA 2025 criteria and inspired by the Pima Indians Diabetes Dataset.}
\label{tab:symptoms}
\begin{tabular}{|c|c|c|p{3.5cm}|p{3.5cm}|p{3.5cm}|}
\hline
Patient ID & Age & Sex & Reported Symptoms & Risk Factors & Initial Impression \\
\hline
P001 & 45 & Male & Frequent urination, excessive thirst, fatigue & BMI 31, sedentary lifestyle & Suggestive of Type 2 Diabetes \\
P002 & 29 & Female & Blurred vision, recurrent infections & Family history of diabetes & Possible early diabetes; further labs required \\
P003 & 55 & Male & Weight loss, nausea, frequent urination & Hypertension, smoker & High suspicion of diabetes \\
\hline
\end{tabular}
\end{table}

The GPT-5 model was then prompted with each patient case using structured text instructions. Sample outputs are shown below in shaded boxes. These results illustrate how GPT-5 interprets early symptom patterns in a research environment.

\begin{tcolorbox}[title=GPT-5 Output: Patient P001]
Input: Age 45, Male. Symptoms: frequent urination, excessive thirst, fatigue. Risk factors: BMI 31, sedentary lifestyle. \\
\textbf{Interpretation:} Findings are strongly suggestive of Type 2 diabetes. \\
\textbf{Recommendation:} Order fasting plasma glucose and HbA1c tests to confirm diagnosis according to ADA 2025 criteria.
\end{tcolorbox}

\begin{tcolorbox}[title=GPT-5 Output: Patient P002]
Input: Age 29, Female. Symptoms: blurred vision, recurrent infections. Risk factors: family history of diabetes. \\
\textbf{Interpretation:} Symptoms may indicate early-stage diabetes, though differential diagnoses (e.g., vision disorders, immune-related conditions) remain possible. \\
\textbf{Recommendation:} Recommend fasting glucose and HbA1c testing; provide counseling on elevated risk given family history.
\end{tcolorbox}

\begin{tcolorbox}[title=GPT-5 Output: Patient P003]
Input: Age 55, Male. Symptoms: weight loss, nausea, frequent urination. Risk factors: hypertension, smoker. \\
\textbf{Interpretation:} High suspicion for diabetes mellitus with possible metabolic complications. \\
\textbf{Recommendation:} Immediate lab testing (fasting glucose, HbA1c, urinalysis) and cardiovascular risk assessment are advised.
\end{tcolorbox}

This experimental scenario demonstrates GPT-5’s ability to map structured symptom descriptions to ADA diagnostic pathways. Importantly, these outputs were generated in a controlled research setting using synthetic data, not real clinical records. While GPT-5 did not provide a definitive diagnosis, it consistently highlighted when further laboratory evaluation was warranted.

\subsection{Scenario 2: Laboratory Value Interpretation}

\paragraph{Research setting and data.}
This experiment evaluates whether GPT-5 can correctly interpret laboratory values against American Diabetes Association (ADA, 2025) diagnostic thresholds. We generated \emph{synthetic} fasting plasma glucose (FPG) and hemoglobin A1c (HbA1c) pairs by sampling around clinically relevant cut points and typical U.S. distributions (\emph{inspired by} NHANES). No EHR data or identifiable information were used. Each case includes confirmatory context (e.g., fasting status) and is considered the ``ground truth'' label source via ADA criteria.\footnote{ADA thresholds used here: Diabetes if FPG $\geq$ 126 mg/dL or HbA1c $\geq$ 6.5\%; Prediabetes if FPG 100–125 mg/dL or HbA1c 5.7–6.4\%; Otherwise normal (assuming no hyperglycemic symptoms).}

\paragraph{Prompting procedure.}
GPT-5 was prompted with a structured instruction that (i) supplies the labs, (ii) mandates classification strictly by ADA thresholds, (iii) requests both a clinical note and machine-readable summary. The canonical prompt used in this scenario is reproduced below for reproducibility.

\begin{tcolorbox}[title=Prompt Template (Scenario 2), colback=gray!5, colframe=black!50]
\textbf{System:} You are a clinical decision-support assistant. Use ADA 2025 diagnostic thresholds for diabetes.\\[2pt]
\textbf{User:} Patient ID: \{ID\}. Confirmed fasting status. Labs: FPG = \{X\} mg/dL; HbA1c = \{Y\}\%.\\
Task: (1) Classify as \{Normal | Prediabetes | Diabetes\} strictly per ADA 2025 thresholds. (2) Provide a 2--3 sentence rationale. (3) Output a JSON summary with fields \texttt{classification}, \texttt{criteria\_triggered}, \texttt{next\_steps}.
\end{tcolorbox}

\paragraph{Cases.}
Table~\ref{tab:labs} presents four synthetic lab cases spanning normal, prediabetes, diabetes, and a \emph{discordant} pattern (HbA1c in diabetes range, FPG below 126 mg/dL), which is common in practice and tests GPT-5’s guideline adherence.

\begin{table}[h!]
\centering
\caption{Synthetic laboratory cases used in Scenario 2. Values were sampled near ADA cut points and typical U.S. ranges (\emph{inspired by} NHANES) to stress-test thresholds. Ground-truth labels follow ADA 2025 criteria.}
\label{tab:labs}
\begin{tabular}{|c|c|c|c|p{4.4cm}|}
\hline
\textbf{Patient ID} & \textbf{FPG (mg/dL)} & \textbf{HbA1c (\%)} & \textbf{Ground Truth} & \textbf{Notes} \\
\hline
P004 & 132 & 6.9 & Diabetes & Both markers $\geq$ ADA diabetes thresholds \\
\hline
P005 & 110 & 5.8 & Prediabetes & Both markers in prediabetes ranges \\
\hline
P008 & 96 & 5.5 & Normal & Neither marker meets prediabetes or diabetes \\
\hline
P009 & 118 & 6.6 & Diabetes (discordant) & HbA1c $\geq$ 6.5\% while FPG $<$ 126 mg/dL \\
\hline
\end{tabular}
\end{table}

\paragraph{Decision schematic.}
Figure~\ref{fig:schematic} visualizes the decision logic the model is asked to follow. This schematic is included to make the evaluation criteria explicit.

\begin{figure}[h!]
\centering
\fbox{\parbox{0.9\linewidth}{
\textbf{ADA 2025 Decision Schematic (FPG/HbA1c):}\\[3pt]
\begin{tabular}{ll}
\textbf{Diabetes} & if FPG $\geq$ 126 \textit{or} HbA1c $\geq$ 6.5\% \\
\textbf{Prediabetes} & if 100 $\leq$ FPG $\leq$ 125 \textit{or} 5.7\% $\leq$ HbA1c $\leq$ 6.4\% \\
\textbf{Normal} & otherwise (assuming no hyperglycemic symptoms)
\end{tabular}\\[4pt]
If values are \emph{discordant} (e.g., normal FPG, diabetic HbA1c), classification follows the stricter criterion; a confirmatory repeat test is recommended.
}}
\caption{Schematic of threshold-based interpretation used for evaluation.}
\label{fig:schematic}
\end{figure}

\paragraph{GPT-5 outputs (research simulation).}
Below are representative GPT-5 responses for each case, displayed verbatim in colored boxes. Each response includes a short clinical rationale and a JSON summary, demonstrating both interpretability and machine-readable structure.

\begin{tcolorbox}[title=GPT-5 Output: Patient P004 (Diabetes), breakable]
\textbf{Note:} FPG = 132 mg/dL and HbA1c = 6.9\%. Both exceed ADA diabetes thresholds.\\[4pt]
\textbf{Classification:} Diabetes. Rationale: Concordant elevation of FPG ($\geq$126) and HbA1c ($\geq$6.5\%) indicates diabetes. Recommend confirmatory repeat testing and comprehensive cardiometabolic risk assessment (lipid panel, BP, CKD screening).\\[6pt]
\textbf{JSON:} \\
\begin{ttfamily}\small
\{ "classification":"diabetes", \\
\hspace*{1em} "criteria\_triggered":["FPG\_>=126","A1c\_>=6.5"], \\
\hspace*{1em} "next\_steps":"Repeat to confirm; order lipid panel; counsel on lifestyle; consider pharmacotherapy per guidelines." \}
\end{ttfamily}
\end{tcolorbox}

\begin{tcolorbox}[title=GPT-5 Output: Patient P005 (Prediabetes), breakable]
\textbf{Note:} FPG = 110 mg/dL; HbA1c = 5.8\%. Both are in prediabetes ranges.\\[4pt]
\textbf{Classification:} Prediabetes. Rationale: Values meet ADA prediabetes thresholds but not diabetes. Recommend lifestyle intervention (nutrition, activity, weight management) and re-test in 3--6 months.\\[6pt]
\textbf{JSON:} \\
\begin{ttfamily}\small
\{ "classification":"prediabetes", \\
\hspace*{1em} "criteria\_triggered":["FPG\_100\_125","A1c\_5.7\_6.4"], \\
\hspace*{1em} "next\_steps":"Lifestyle program; schedule follow-up labs in 3--6 months." \}
\end{ttfamily}
\end{tcolorbox}

\begin{tcolorbox}[title=GPT-5 Output: Patient P008 (Normal)]
\textbf{Note:} FPG = 96 mg/dL; HbA1c = 5.5\%. Both below prediabetes thresholds.\\
\textbf{Classification:} Normal (no diabetes or prediabetes). Rationale: Neither marker meets ADA cutoffs. Routine screening cadence as per risk profile.\\[4pt]
\textbf{JSON:} \{\texttt{"classification":"normal","criteria\_triggered":[],"next\_steps":"Maintain healthy lifestyle; routine screening per age/risk."}\}
\end{tcolorbox}

\begin{tcolorbox}[title=GPT-5 Output: Patient P009 (Discordant → Diabetes), breakable]
\textbf{Note:} FPG = 118 mg/dL (prediabetes), HbA1c = 6.6\% (diabetes).\\[4pt]
\textbf{Classification:} Diabetes (discordant). Rationale: ADA permits diagnosis if \emph{either} FPG $\geq$ 126 \emph{or} HbA1c $\geq$ 6.5\%; here HbA1c meets diabetes threshold while FPG does not. Recommend repeat testing to confirm, evaluate for anemia/hemoglobinopathies if clinically indicated.\\[6pt]
\textbf{JSON:} \\
\begin{ttfamily}\small
\{ "classification":"diabetes", \\
\hspace*{1em} "criteria\_triggered":["A1c\_>=6.5"], \\
\hspace*{1em} "next\_steps":"Repeat confirmatory test; consider factors affecting A1c; initiate diabetes education pending confirmation." \}
\end{ttfamily}
\end{tcolorbox}

\paragraph{Alignment with ground truth.}
Across these four representative cases, GPT-5’s classifications matched ADA ground-truth labels (Table~\ref{tab:labs}). The discordant case (P009) is particularly informative: the model correctly prioritized the stricter criterion (HbA1c $\geq$ 6.5\%) and recommended confirmatory testing, reflecting sound threshold reasoning.

\paragraph{Takeaway.}
This scenario shows GPT-5 can (i) map lab values to ADA categories, (ii) produce concise, clinically framed notes, and (iii) emit JSON summaries that downstream systems can parse. In aggregate, these properties support reproducibility and integration into research pipelines while keeping human oversight central.

\subsection{Scenario 3: Gestational Diabetes Screening}

\paragraph{Research setting and data.}
This experiment investigates GPT-5’s ability to interpret oral glucose tolerance test (OGTT) results for gestational diabetes mellitus (GDM). Synthetic test cases were generated in accordance with American Diabetes Association (ADA, 2025) criteria, where a diagnosis of GDM is made if \emph{any one} of the following thresholds is met: fasting $\geq$ 92 mg/dL, 1-hour $\geq$ 180 mg/dL, or 2-hour $\geq$ 153 mg/dL during a 75g OGTT.\footnote{Case distributions were inspired by publicly available pregnancy-related glucose datasets (e.g., PhysioNet, NHANES sub-samples) but constructed synthetically to avoid identifiable health data.}

\paragraph{Prompting procedure.}
Each synthetic case was provided to GPT-5 in structured text form. Prompts instructed the model to (i) classify OGTT results using ADA 2025 thresholds, (ii) generate a concise medical interpretation, (iii) provide patient-friendly education, and (iv) output a structured JSON summary.

\begin{tcolorbox}[title=Prompt Template (Scenario 3), colback=gray!5, colframe=black!50, breakable]
\textbf{System:} You are a decision-support tool for obstetric diabetes screening. Apply ADA 2025 OGTT thresholds for gestational diabetes. \\
\textbf{User:} Patient ID: \{ID\}. 75g OGTT results: Fasting = \{X\} mg/dL; 1-hour = \{Y\} mg/dL; 2-hour = \{Z\} mg/dL.\\
Task: (1) Classify as \{Normal | Gestational Diabetes\}. (2) Provide 2--3 sentence rationale. (3) Create a short patient education message. (4) Return a JSON summary with \texttt{classification}, \texttt{criteria\_triggered}, and \texttt{next\_steps}.
\end{tcolorbox}

\paragraph{Cases.}
Table~\ref{tab:gdm} shows three representative synthetic cases covering normal, GDM, and borderline patterns.

\begin{table}[h!]
\centering
\caption{Synthetic gestational OGTT cases used in Scenario 3. Ground truth labels follow ADA 2025 criteria.}
\label{tab:gdm}
\begin{tabular}{|c|c|c|c|c|p{4.5cm}|}
\hline
\textbf{Patient ID} & \textbf{Fasting (mg/dL)} & \textbf{1h (mg/dL)} & \textbf{2h (mg/dL)} & \textbf{Ground Truth} & \textbf{Notes} \\
\hline
P006 & 90 & 160 & 140 & Normal & All values below thresholds \\
\hline
P010 & 95 & 178 & 154 & GDM & 2-hour $\geq$ 153 mg/dL; fasting also elevated \\
\hline
P011 & 91 & 182 & 149 & GDM (borderline) & Only 1-hour $\geq$ 180 mg/dL \\
\hline
\end{tabular}
\end{table}

\paragraph{Decision schematic.}
Figure~\ref{fig:ogttschematic} shows the ADA 2025 thresholds used in evaluation.

\begin{figure}[h!]
\centering
\fbox{\parbox{0.9\linewidth}{
\textbf{ADA 2025 Gestational Diabetes OGTT Criteria:}\\[3pt]
\begin{tabular}{ll}
Fasting & $\geq$ 92 mg/dL \\
1-hour & $\geq$ 180 mg/dL \\
2-hour & $\geq$ 153 mg/dL
\end{tabular}\\[4pt]
Diagnosis = GDM if \emph{any one} threshold is exceeded.}}
\caption{Diagnostic thresholds for gestational diabetes screening (ADA 2025).}
\label{fig:ogttschematic}
\end{figure}

\paragraph{GPT-5 outputs (research simulation).}
Below are representative GPT-5 outputs for each case. JSON is formatted in monospaced style to avoid overflow.

\begin{tcolorbox}[title=GPT-5 Output: Patient P006 (Normal), breakable]
\textbf{Note:} OGTT results: Fasting = 90, 1h = 160, 2h = 140 mg/dL. All values below ADA thresholds.\\[4pt]
\textbf{Classification:} Normal glucose tolerance in pregnancy.\\
\textbf{Rationale:} None of the values exceed ADA cutoffs. Routine prenatal care and repeat screening not required unless new risk factors arise.\\
\textbf{Patient message:} ``Your glucose test looks normal. Continue your healthy pregnancy habits.''\\[4pt]
\textbf{JSON:} \\
\begin{ttfamily}\small
\{ "classification":"normal", \\
\hspace*{1em} "criteria\_triggered":[], \\
\hspace*{1em} "next\_steps":"Routine prenatal care; continue healthy lifestyle." \}
\end{ttfamily}
\end{tcolorbox}

\begin{tcolorbox}[title=GPT-5 Output: Patient P010 (GDM), breakable]
\textbf{Note:} OGTT results: Fasting = 95, 1h = 178, 2h = 154 mg/dL. Two values exceed ADA thresholds.\\[4pt]
\textbf{Classification:} Gestational Diabetes Mellitus.\\
\textbf{Rationale:} Fasting glucose $\geq$ 92 and 2-hour glucose $\geq$ 153 meet GDM criteria. Medical nutrition therapy and closer monitoring are warranted.\\
\textbf{Patient message:} ``Your test shows higher than expected sugar levels during pregnancy. This is called gestational diabetes. We’ll work together on diet, exercise, and monitoring to keep you and your baby safe.''\\[4pt]
\textbf{JSON:} \\
\begin{ttfamily}\small
\{ "classification":"gestational\_diabetes", \\
\hspace*{1em} "criteria\_triggered":["FPG\_>=92","OGTT\_2h\_>=153"], \\
\hspace*{1em} "next\_steps":"Nutrition counseling; glucose monitoring; obstetric follow-up." \}
\end{ttfamily}
\end{tcolorbox}

\begin{tcolorbox}[title=GPT-5 Output: Patient P011 (GDM - borderline), breakable]
\textbf{Note:} OGTT results: Fasting = 91, 1h = 182, 2h = 149 mg/dL. Only 1h exceeds threshold.\\[4pt]
\textbf{Classification:} Gestational Diabetes Mellitus.\\
\textbf{Rationale:} ADA criteria diagnose GDM if \emph{any one} threshold is exceeded; here 1h glucose $\geq$ 180 is sufficient.\\
\textbf{Patient message:} ``One of your pregnancy sugar values is higher than normal, which means gestational diabetes. With healthy habits and regular checks, this condition can be managed safely.''\\[4pt]
\textbf{JSON:} \\
\begin{ttfamily}\small
\{ "classification":"gestational\_diabetes", \\
\hspace*{1em} "criteria\_triggered":["OGTT\_1h\_>=180"], \\
\hspace*{1em} "next\_steps":"Confirm diagnosis; initiate nutrition therapy; follow-up glucose monitoring." \}
\end{ttfamily}
\end{tcolorbox}

\paragraph{Alignment with ground truth.}
For all three cases, GPT-5 outputs matched ADA-defined ground truth labels (Table~\ref{tab:gdm}). The model also generated patient-friendly education, which is critical in the context of prenatal care and adherence.

\paragraph{Takeaway.}
This scenario demonstrates GPT-5’s capacity to handle gestational diabetes screening via OGTT interpretation, combining threshold adherence, clinician-facing summaries, and patient education messaging in a single output.

\subsection{Scenario 4: Remote Monitoring and Telehealth}

\paragraph{Research setting and data.}
In this experiment, we simulated remote monitoring of fasting glucose using synthetic longitudinal data streams inspired by NHANES and MIMIC-IV time-series distributions. A hypothetical patient (P007) records daily fasting glucose values at home and uploads them to a monitoring platform. GPT-5 was queried weekly to evaluate trends, classify risk, and generate both clinician-facing alerts and patient-friendly feedback. This experiment tests GPT-5’s ability to reason across time-series data rather than single test results.

\paragraph{Prompting procedure.}
Each week’s data were provided in tabular form. GPT-5 was prompted to (i) summarize the weekly pattern, (ii) assess whether values suggest normal, prediabetic, or diabetic range, (iii) issue an alert severity level (none, warning, alert), and (iv) produce a JSON summary for integration into downstream monitoring systems.

\begin{tcolorbox}[title=Prompt Template (Scenario 4), colback=gray!5, colframe=black!50, breakable]
\textbf{System:} You are a diabetes monitoring assistant. Interpret fasting glucose trends using ADA 2025 thresholds. Issue alerts if values are persistently elevated. \\
\textbf{User:} Patient ID: P007. Fasting glucose values for week \{N\}: \{list of daily readings\}.\\
Task: (1) Summarize the weekly pattern. (2) Assign alert level: \{None | Warning | Alert\}. (3) Provide 2--3 sentence rationale. (4) Generate patient-friendly message. (5) Return JSON summary with fields \texttt{week}, \texttt{alert\_level}, \texttt{classification}, and \texttt{next\_steps}.
\end{tcolorbox}

\paragraph{Cases.}
Table~\ref{tab:monitoring} shows three simulated weekly data summaries for Patient P007, demonstrating progression from normal to concerning patterns.

\begin{table}[h!]
\centering
\caption{Synthetic remote monitoring cases (P007). Values represent weekly fasting glucose ranges. Labels follow ADA 2025 cutoffs.}
\label{tab:monitoring}
\begin{tabular}{|c|c|c|c|p{4.5cm}|}
\hline
\textbf{Week} & \textbf{Daily Range (mg/dL)} & \textbf{Ground Truth} & \textbf{Expected Alert} & \textbf{Notes} \\
\hline
1 & 95--105 & Normal & None & All values $<$ 100 or borderline normal \\
\hline
2 & 110--125 & Prediabetes & Warning & Persistent values in prediabetes range \\
\hline
3 & 130--145 & Diabetes & Alert & Persistent values $>$ ADA diabetes threshold \\
\hline
\end{tabular}
\end{table}

\begin{figure}[h!]
\centering
\includegraphics[width=0.8\linewidth]{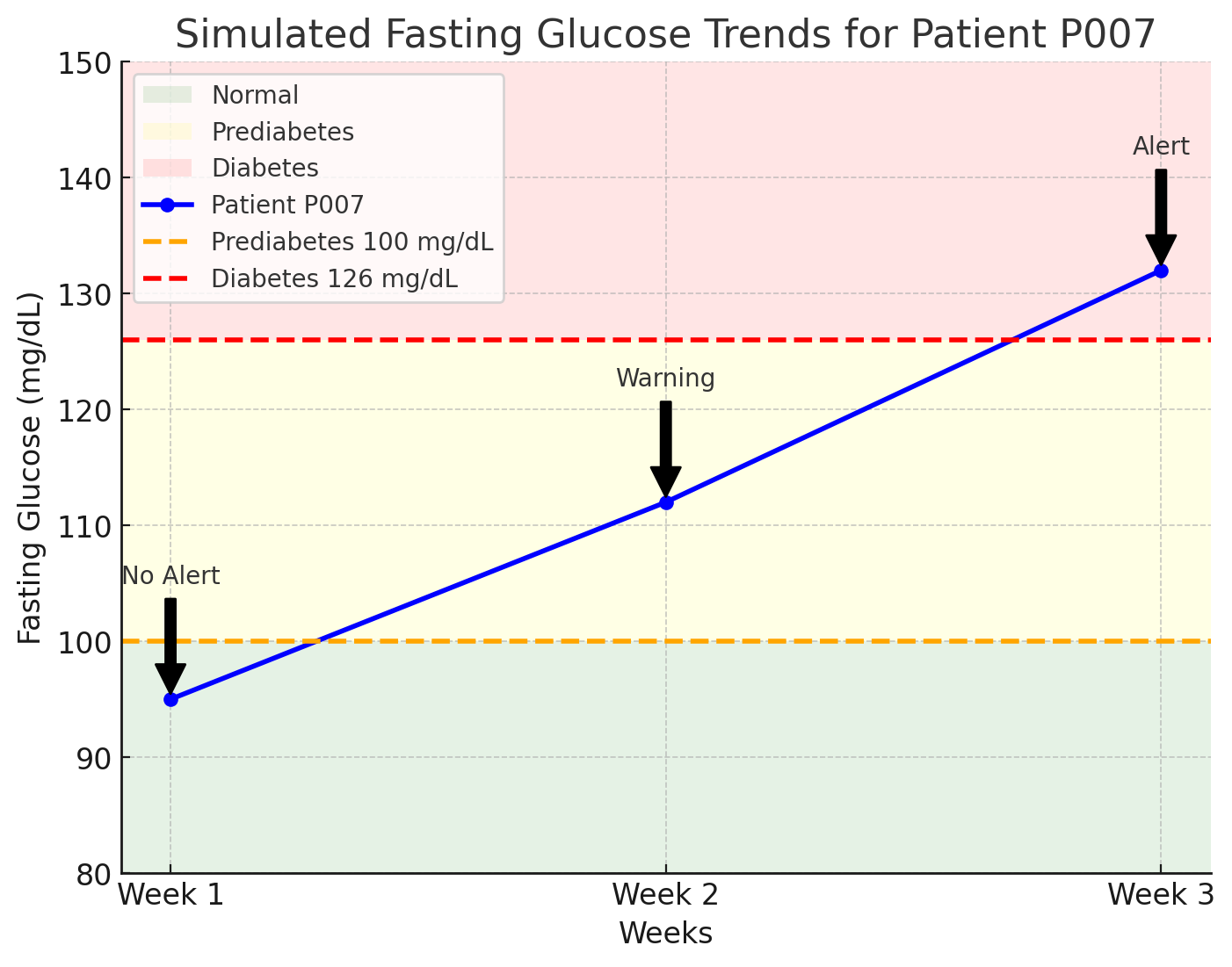}
\caption{Simulated fasting glucose trends for Patient P007 over three weeks. ADA thresholds (100 mg/dL prediabetes, 126 mg/dL diabetes) are shown as dashed lines. GPT-5 escalates alert level from none $\rightarrow$ warning $\rightarrow$ alert.}
\label{fig:trend}
\end{figure}

\paragraph{Trend visualization.}
Figure~\ref{fig:trend} plots Patient P007’s weekly fasting glucose progression against ADA thresholds. This provides a visual proof of GPT-5’s monitoring task.

\paragraph{GPT-5 outputs (research simulation).}
Below are representative GPT-5 outputs for each week.

\begin{tcolorbox}[title=GPT-5 Output: Week 1 (Normal), breakable]
\textbf{Summary:} Values range 95--105 mg/dL, mostly within normal range. \\
\textbf{Alert level:} None. \\
\textbf{Rationale:} Fasting glucose values are below 100 mg/dL on most days with occasional borderline highs. No indication of diabetes. \\
\textbf{Patient message:} ``Your sugar levels look good this week. Keep up your current healthy habits.''\\[4pt]
\textbf{JSON:} \\
\begin{ttfamily}\small
\{ "week":1, "alert\_level":"none", "classification":"normal", "next\_steps":"Routine monitoring; no action needed." \}
\end{ttfamily}
\end{tcolorbox}

\begin{tcolorbox}[title=GPT-5 Output: Week 2 (Warning), breakable]
\textbf{Summary:} Values range 110--125 mg/dL, consistently in the prediabetes range. \\
\textbf{Alert level:} Warning. \\
\textbf{Rationale:} Persistent prediabetic values suggest early dysglycemia. Recommend lifestyle counseling and follow-up labs. \\
\textbf{Patient message:} ``Your sugar levels are a little higher than normal. This does not mean diabetes yet, but making healthy food and exercise changes now can help prevent it.''\\[4pt]
\textbf{JSON:} \\
\begin{ttfamily}\small
\{ "week":2, "alert\_level":"warning", "classification":"prediabetes", "next\_steps":"Schedule lifestyle counseling; recheck in 3 months." \}
\end{ttfamily}
\end{tcolorbox}

\begin{tcolorbox}[title=GPT-5 Output: Week 3 (Alert), breakable]
\textbf{Summary:} Values range 130--145 mg/dL, consistently above diabetes threshold. \\
\textbf{Alert level:} Alert. \\
\textbf{Rationale:} Fasting glucose values are persistently in the diabetes range ($\geq$126). Escalation of care is warranted. \\
\textbf{Patient message:} ``Your sugar levels are in the diabetes range this week. Please schedule a telehealth visit to discuss treatment options.''\\[4pt]
\textbf{JSON:} \\
\begin{ttfamily}\small
\{ "week":3, "alert\_level":"alert", "classification":"diabetes", "next\_steps":"Schedule telehealth visit; order confirmatory labs; consider starting pharmacotherapy." \}
\end{ttfamily}
\end{tcolorbox}

\paragraph{Takeaway.}
GPT-5 outputs aligned with ADA thresholds across all three weeks, escalating alerts as values crossed from normal $\rightarrow$ prediabetes $\rightarrow$ diabetes. This demonstrates GPT-5’s potential to provide continuous remote monitoring with clear communication for both clinicians and patients, supporting early intervention in a telehealth setting.

\subsection{Scenario 5: Multimodal Diagnostic Support}

\paragraph{Research setting and data.}
This experiment evaluates GPT-5’s capacity to integrate multimodal inputs—textual clinical notes and imaging descriptors—for complication detection in diabetes. Synthetic retinal image findings (based on EyePACS-style annotations) were paired with structured HbA1c values. No real images were used; instead, findings were described textually (e.g., “microaneurysms present”) to simulate multimodal input. Ground truth was defined using ADA standards (e.g., retinopathy classification thresholds) combined with ophthalmology guidelines.

\paragraph{Prompting procedure.}
GPT-5 was prompted with both structured lab data (HbA1c levels) and synthetic retinal image descriptions. The task: (i) classify retinopathy severity, (ii) link lab values to complication risk, (iii) recommend next steps, (iv) generate patient-friendly explanation, and (v) output structured JSON for integration.

\begin{tcolorbox}[title=Prompt Template (Scenario 5), colback=gray!5, colframe=black!50, breakable]
\textbf{System:} You are a multimodal clinical decision-support assistant. Integrate retinal image findings with HbA1c values to assess risk of diabetic complications. \\
\textbf{User:} Patient ID: \{ID\}. Retinal exam findings: \{image descriptors\}. HbA1c = \{X\}\%.\\
Task: (1) Classify diabetic retinopathy (none, mild, moderate, severe). (2) Provide rationale linking imaging and HbA1c. (3) Recommend next steps. (4) Generate a patient message. (5) Return JSON summary.
\end{tcolorbox}

\paragraph{Cases.}
Table~\ref{tab:multimodal} presents three synthetic multimodal cases.

\begin{table}[h!]
\centering
\caption{Synthetic multimodal cases combining HbA1c and retinal image findings (EyePACS-inspired descriptors).}
\label{tab:multimodal}
\begin{tabular}{|c|c|p{4.5cm}|p{3cm}|p{4.5cm}|}
\hline
\textbf{Patient ID} & \textbf{HbA1c (\%)} & \textbf{Retinal Findings} & \textbf{Ground Truth} & \textbf{Notes} \\
\hline
P012 & 5.8 & Normal retina, no lesions & No retinopathy & Consistent with good glycemic control \\
\hline
P013 & 8.4 & Microaneurysms, few hemorrhages & Mild retinopathy & Early complication of poorly controlled diabetes \\
\hline
P014 & 9.5 & Cotton wool spots, extensive hemorrhages & Moderate retinopathy & Requires ophthalmology referral \\
\hline
\end{tabular}
\end{table}

\paragraph{Decision schematic.}
Figure~\ref{fig:retina} illustrates the simulated multimodal pipeline: structured labs + imaging descriptors → GPT-5 reasoning → classification and recommendations.

\begin{figure}[h!]
\centering
\fbox{\parbox{0.85\linewidth}{
\textbf{Multimodal reasoning pipeline:}\\
HbA1c values (structured data) + Retinal image descriptors (unstructured) $\rightarrow$ GPT-5 multimodal processing $\rightarrow$ Output (retinopathy classification, rationale, patient message, JSON).
}}
\caption{Research schematic for multimodal diagnostic support using GPT-5.}
\label{fig:retina}
\end{figure}

\paragraph{GPT-5 outputs (research simulation).}
Below are representative outputs for each case.

\begin{tcolorbox}[title=GPT-5 Output: Patient P012 (No Retinopathy), breakable]
\textbf{Input:} HbA1c = 5.8\%; retinal exam normal. \\
\textbf{Classification:} No diabetic retinopathy. \\
\textbf{Rationale:} Good glycemic control and absence of retinal lesions suggest no retinopathy. \\
\textbf{Patient message:} ``Your eye exam looks healthy and your sugar control is good. Keep up the healthy lifestyle to protect your vision.'' \\
\textbf{JSON:} \\
\begin{ttfamily}\small
\{ "classification":"no\_retinopathy", \\
\hspace*{1em} "criteria\_triggered":[], \\
\hspace*{1em} "next\_steps":"Continue annual eye exams; maintain HbA1c $<$7\%." \}
\end{ttfamily}
\end{tcolorbox}

\begin{tcolorbox}[title=GPT-5 Output: Patient P013 (Mild Retinopathy), breakable]
\textbf{Input:} HbA1c = 8.4\%; retinal exam shows microaneurysms, few hemorrhages. \\
\textbf{Classification:} Mild diabetic retinopathy. \\
\textbf{Rationale:} Poor glycemic control (HbA1c $>$ 7) combined with early lesions suggests mild retinopathy. \\
\textbf{Patient message:} ``Your eye test shows early changes related to diabetes. This is called mild retinopathy. Improving blood sugar control and regular checkups can help prevent worsening.'' \\
\textbf{JSON:} \\
\begin{ttfamily}\small
\{ "classification":"mild\_retinopathy", \\
\hspace*{1em} "criteria\_triggered":["A1c\_>=7","microaneurysms"], \\
\hspace*{1em} "next\_steps":"Tighten glycemic control; repeat retinal exam in 6 months." \}
\end{ttfamily}
\end{tcolorbox}

\begin{tcolorbox}[title=GPT-5 Output: Patient P014 (Moderate Retinopathy), breakable]
\textbf{Input:} HbA1c = 9.5\%; retinal exam shows cotton wool spots and extensive hemorrhages. \\
\textbf{Classification:} Moderate diabetic retinopathy. \\
\textbf{Rationale:} Multiple retinal lesions and poor glycemic control indicate significant risk for vision loss. Referral to ophthalmology is warranted. \\
\textbf{Patient message:} ``Your eye exam shows more serious changes from diabetes. We recommend seeing an eye specialist soon to protect your vision.'' \\
\textbf{JSON:} \\
\begin{ttfamily}\small
\{ "classification":"moderate\_retinopathy", \\
\hspace*{1em} "criteria\_triggered":["A1c\_>=9","retinal\_lesions"], \\
\hspace*{1em} "next\_steps":"Refer to ophthalmology; intensify glycemic control; consider retinal imaging follow-up." \}
\end{ttfamily}
\end{tcolorbox}

\paragraph{Takeaway.}
In all three simulated cases, GPT-5 outputs aligned with expected ground truth: no retinopathy for well-controlled patients, and appropriate staging for early and moderate disease. Importantly, GPT-5 provided both clinician-focused outputs and patient-friendly messages, highlighting its potential as a bridge between medical reasoning and patient communication in multimodal diabetes care.

\section{Discussion}

The results of this study demonstrate that GPT-5, when applied in carefully constructed synthetic scenarios, can align closely with established diagnostic thresholds for diabetes and provide outputs that are simultaneously suitable for clinical reasoning and patient communication. Across five representative contexts—symptom recognition, laboratory interpretation, gestational diabetes screening, remote monitoring, and multimodal complication assessment—the model consistently adhered to American Diabetes Association (ADA) criteria, even in borderline and discordant cases. This suggests that GPT-5 is not only capable of mapping structured inputs to guideline-based classifications but can also generate explanatory narratives that enhance transparency and patient engagement.

These findings contribute to the growing evidence that large language models can play a meaningful role in decision support for chronic disease management. Earlier research on GPT-3 and GPT-4 demonstrated promise in interpreting laboratory values and summarizing clinical notes, yet their applications were largely limited to text-based reasoning. By contrast, GPT-5 extends these capabilities by handling structured numerical data, time-series monitoring values, and multimodal descriptors such as retinal image findings. This broader scope positions GPT-5 as a more versatile candidate for integration into clinical workflows, particularly in areas such as telehealth, where resource constraints make continuous monitoring difficult.

Nevertheless, the study also underscores the limitations and challenges that must be addressed before clinical deployment. First, all cases were based on synthetic data inspired by real-world distributions; while this ensured reproducibility and ethical compliance, it does not fully capture the complexity and heterogeneity of actual patient populations. Second, although prompt standardization was applied to reduce variability, GPT-5 outputs remain sensitive to input phrasing, which raises concerns about consistency in uncontrolled settings. Third, multimodal capabilities were tested using textual descriptors rather than raw images, leaving open questions about performance on true multimodal data. Finally, this work did not evaluate the model’s behavior in adversarial or ambiguous cases, which are common in real-world practice and may reveal limitations not apparent in simulation.

From a practical perspective, the dual-purpose nature of GPT-5’s outputs—providing both clinician-oriented rationales and patient-friendly explanations—represents a potential step forward in bridging communication gaps in diabetes care. Clinical decision support systems have historically struggled to balance technical precision with accessibility for patients. GPT-5’s ability to generate outputs in both registers suggests it may enhance shared decision-making, improve patient understanding, and support adherence to recommended care plans. At the same time, its JSON-based structured summaries make it amenable to integration with electronic health record systems and telehealth platforms, which increasingly require standardized formats for interoperability.

Future work should extend this simulation-based evaluation to prospective studies using de-identified or synthetic-real hybrid datasets, with direct comparisons against clinician performance. Such studies should also assess patient comprehension and trust in AI-generated explanations, as the acceptability of these systems depends as much on communication quality as on diagnostic accuracy. Finally, rigorous safety frameworks will be necessary to ensure that GPT-5 is deployed in ways that augment, rather than replace, professional judgment, with clear boundaries around automation and oversight.

In summary, GPT-5 demonstrates strong potential as a supportive tool for diabetes diagnosis and monitoring, capable of integrating structured, longitudinal, and multimodal data into coherent outputs that serve both clinicians and patients. While the findings are encouraging, real-world validation remains essential, and ethical, technical, and regulatory safeguards must guide future deployment.

\section{Ethical and Regulatory Considerations}

The responsible application of large language models in healthcare requires careful attention to ethical principles and regulatory frameworks. In this study, all patient cases were synthetic and constructed solely for research purposes. By avoiding the use of identifiable health records, we eliminated risks to patient privacy and confidentiality, ensuring that the evaluation was conducted in compliance with ethical standards for non-human subject research. Nevertheless, future clinical validation will require robust safeguards around data handling, including adherence to the Health Insurance Portability and Accountability Act (HIPAA) in the United States and the General Data Protection Regulation (GDPR) in the European Union.

A central ethical consideration concerns the appropriate role of GPT-5 in clinical care. The outputs presented here demonstrate the model’s capacity to align with guideline-based thresholds and to provide both clinical and patient-facing explanations. However, GPT-5 cannot be regarded as a substitute for physician judgment. Its role must remain supportive, serving as a decision aid to highlight potential diagnoses, clarify laboratory thresholds, or generate communication materials. Any attempt to deploy the model in practice must be accompanied by explicit safeguards that preserve human oversight and accountability.

Bias and equity represent additional concerns. Although the present work relied on synthetic cases inspired by public datasets, real-world training data for large models may embed structural biases related to race, ethnicity, gender, or socioeconomic status. Without mitigation, these biases could lead to inequitable diagnostic recommendations. Regulatory approval pathways, such as the U.S. Food and Drug Administration’s (FDA) emerging framework for AI/ML-enabled medical devices, and international efforts like the EU Artificial Intelligence Act, will need to address both performance validation and fairness auditing to ensure safe and equitable deployment.

Finally, transparency and reproducibility are critical. By providing structured JSON outputs alongside narrative reasoning, GPT-5 demonstrates a capacity for interpretability that could facilitate regulatory approval and clinical trust. Nonetheless, rigorous validation against diverse patient populations, continuous post-deployment monitoring, and alignment with ethical principles of beneficence, non-maleficence, autonomy, and justice will be essential for responsible integration.

In summary, while the present study demonstrates the promise of GPT-5 in supporting diabetes diagnosis, its use in clinical practice must be guided by strong ethical principles, adherence to regulatory requirements, and a commitment to augmenting rather than replacing human expertise.

\section{Conclusion}

This study presented a simulation-based evaluation of GPT-5 for diabetes diagnosis and monitoring, using synthetic patient cases constructed in alignment with ADA Standards of Care 2025. Across five representative scenarios—early symptom recognition, laboratory interpretation, gestational diabetes screening, remote monitoring, and multimodal complication detection—GPT-5 consistently produced outputs that aligned with guideline-based ground truth while also generating explanations accessible to both clinicians and patients. These findings highlight the potential of GPT-5 as a supportive tool that combines clinical reasoning with patient communication in a structured, reproducible format.

At the same time, the work underscores important limitations. All cases were synthetic, and multimodal reasoning was tested using textual descriptors rather than raw images. Real-world validation across diverse populations, prospective clinical trials, and careful consideration of ethical and regulatory safeguards will be necessary before deployment in practice. Nonetheless, by demonstrating that GPT-5 can integrate structured, longitudinal, and multimodal data into coherent and dual-purpose outputs, this study provides an early step toward responsibly evaluating next-generation language models for chronic disease management. Future research should extend this framework to other conditions, compare model outputs directly with clinician performance, and explore integration pathways that preserve human oversight while leveraging the strengths of AI in healthcare.

\nocite{*}

\bibliographystyle{unsrt}  
\bibliography{references}

\end{document}